# Computing Posterior Probabilities of Structural Features in Bayesian Networks

**Jin Tian and Ru He**
Department of Computer Science
Iowa State University
Ames, IA 50011
{*jtian, rhe*}@cs.iastate.edu

## Abstract

We study the problem of learning Bayesian network structures from data. Koivisto and Sood (2004) and Koivisto (2006) presented algorithms that can compute the exact marginal posterior probability of a subnetwork, e.g., a single edge, in $O(n2^n)$ time and the posterior probabilities for all $n(n-1)$ potential edges in $O(n2^n)$ total time, assuming that the number of parents per node or the indegree is bounded by a constant. One main drawback of their algorithms is the requirement of a special structure prior that is non uniform and does not respect Markov equivalence. In this paper, we develop an algorithm that can compute the exact posterior probability of a subnetwork in $O(3^n)$ time and the posterior probabilities for all $n(n-1)$ potential edges in $O(n3^n)$ total time. Our algorithm also assumes a bounded indegree but allows general structure priors. We demonstrate the applicability of the algorithm on several data sets with up to 20 variables.

## 1 Introduction

Bayesian networks are being widely used for probabilistic inference and causal modeling [Pearl, 2000, Spirtes *et al.*, 2001]. One major challenge is to learn the structures of Bayesian networks from data. In the Bayesian approach, we provide a prior probability distribution over the space of possible Bayesian networks and then computer the posterior distributions $P(G|D)$ of the network structure $G$ given data $D$. We can then compute the posterior probability of any hypothesis of interest by averaging over all possible networks. In many applications we are interested in structural features. For example, in causal discovery, we are interested in the causal relations among variables, represented by the edges in the network structure [Heckerman *et al.*, 1999].

The number of possible network structures is superexponential $O(n!2^{n(n-1)/2})$ in the number of variables $n$. For example, there are about $10^4$ directed acyclic graphs (DAGs) on 5 nodes, and $10^{18}$ DAGs on 10 nodes. As a result, it is impractical to sum over all possible structures unless for very small networks (less than 8 variables). One solution is to compute approximate posterior probabilities. Madigan and York (1995) used Markov chain Monte Carlo (MCMC) algorithm in the space of network structures. Friedman and Koller (2003) developed a MCMC procedure in the space of node orderings which was shown to be more efficient than MCMC in the space of DAGs. One problem to the MCMC approach is that there is no guarantee on the quality of the approximation in finite runs.

Recently, a dynamic programming (DP) algorithm was developed that can compute the exact marginal posterior probabilities of any subnetwork (e.g., an edge) in $O(n2^n)$ time [Koivisto and Sood, 2004] and the exact posterior probabilities for all $n(n-1)$ potential edges in $O(n2^n)$ total time [Koivisto, 2006], assuming that the *indegree*, i.e., the number of parents of each node, is bounded by a constant. One main drawback of the DP algorithm and the order MCMC algorithm is that they both require a special form of the structure prior $P(G)$. The resulting prior $P(G)$ is non uniform, and does not respect Markov equivalence [Friedman and Koller, 2003, Koivisto and Sood, 2004]. Therefore the computed posterior probabilities could be biased. MCMC algorithms have been developed that try to fix this structure prior problem [Eaton and Murphy, 2007, Ellis and Wong, 2008].

Inspired by the DP algorithm in [Koivisto and Sood, 2004, Koivisto, 2006], we have developed an algorithm for computing the exact posterior probabilities of structural features that does not require a special prior $P(G)$ other than the stan-dard structure modularity (see Eq. (4)). Assuming a bounded indegree, our algorithm can compute the exact marginal posterior probabilities of any subnetwork in $O(3^n)$ time, and the posterior probabilities for all $n(n-1)$ potential edges in $O(n3^n)$ total time. The memory requirement of our algorithm is about the same $O(n2^n)$ as the DP algorithm. We have demonstrated our algorithm on data sets with up to 20 variables. The main advantage of our algorithm is that it can use very general structure prior $P(G)$ that can simply be left as uniform and can satisfy Markov equivalence requirement. We acknowledge here that our algorithm was inspired by and used many techniques in [Koivisto and Sood, 2004, Koivisto, 2006]. Their algorithm is based on summing over all possible total orders (leading to the bias on prior $P(G)$ that graphs consistent with more orders are favored). Our algorithm directly sums over all possible DAG structures by exploiting *sinks*, nodes that have no outgoing edges, and *roots*, nodes that have no parents, and as a result the computations involved are more complicated. We note that the dynamic programming techniques have also been used to learn the optimal Bayesian networks in [Singh and Moore, 2005, Silander and Myllymaki, 2006].

The rest of the paper is organized as follows. In Section 2 we briefly review the Bayesian approach to learn Bayesian networks from data. In Section 3 we present our algorithm for computing the posterior probability of a single edge and in Section 4 we present our algorithm for computing the posterior probabilities of all potential edges simultaneously. We empirically demonstrate the capability of our algorithm in Section 5 and discuss its potential applications in Section 6.

## 2 Bayesian Learning of Bayesian Networks

A Bayesian network is a DAG $G$ that encodes a joint probability distribution over a set $X = \{X_1, \ldots, X_n\}$ of random variables with each node of the graph representing a variable in $X$. For convenience we will typically work on the index set $V = \{1, \ldots, n\}$ and represent a variable $X_i$ by its index $i$. We use $X_{Pa_i} \subseteq X$ to represent the set of parents of $X_i$ in a DAG $G$ and use $Pa_i \subseteq V$ to represent the corresponding index set.

Assume we are given a training data set $D = \{x^1, x^2, \ldots, x^N\}$, where each $x^i$ is a particular instantiation over the set of variables $X$. We only consider situations where the data are complete, that is, every variable in $X$ is assigned a value. In the Bayesian approach to learn Bayesian networks from the training data $D$, we compute the posterior probability of a network $G$ as

$$P(G|D) = \frac{P(D|G)P(G)}{P(D)}. \quad (1)$$

We can then compute the posterior probability of any hypothesis of interest by averaging over all possible networks. In this paper, we are interested in computing the posteriors of structural features. Let $f$ be a structural feature represented by an indicator function such that $f(G)$ is 1 if the feature is present in $G$ and 0 otherwise. We have

$$P(f|D) = \sum_G f(G)P(G|D). \quad (2)$$

Assuming global and local parameter independence, and parameter modularity, $P(D|G)$ can be decomposed into a product of local marginal likelihood often called local scores as [Cooper and Herskovits, 1992, Heckerman *et al.*, 1995]

$$P(D|G) = \prod_{i=1}^n P(x_i|x_{Pa_i} : D) \equiv \prod_{i=1}^n score_i(Pa_i : D), \quad (3)$$

where, with appropriate parameter priors, $score_i(Pa_i : D)$ has a closed form solution. In this paper we will assume that these local scores can be computed efficiently from data.

For the prior $P(G)$ over all possible DAG structures we will assume *structure modularity*: [Friedman and Koller, 2003]

$$P(G) = \prod_{i=1}^n Q_i(Pa_i). \quad (4)$$

In this paper we consider *modular features*:

$$f(G) = \prod_{i=1}^n f_i(Pa_i), \quad (5)$$

where $f_i(Pa_i)$ is an indicator function with values either 0 or 1. For example, an edge $j \to i$ can be represented by setting $f_i(Pa_i) = 1$ if and only if $j \in Pa_i$ and setting $f_l(Pa_l) = 1$ for all $l \neq i$. In this paper, we are interested in computing the posterior $P(f|D)$ of the feature, which can be obtained by computing the joint probability $P(f, D)$ as

$$P(f, D) = \sum_G f(G)P(D|G)P(G) \quad (6)$$

$$= \sum_G \prod_{i=1}^n f_i(Pa_i)Q_i(Pa_i)score_i(Pa_i : D)$$

$$= \sum_G \prod_{i=1}^n B_i(Pa_i), \quad (7)$$

where for all $Pa_i \subseteq V - \{i\}$ we define

$$B_i(Pa_i) \equiv f_i(Pa_i)Q_i(Pa_i)score_i(Pa_i : D). \quad (8)$$

It is clear from Eq. (6) that if we set all features $f_i(Pa_i)$ to be constant 1 then we have $P(f = 1, D) = P(D)$. Therefore we can compute the posterior $P(f|D)$ if we know how to compute the joint $P(f, D)$. In the next section, we show how the summation in Eq. (7) can be done by dynamic programming in time complexity $O(3^n)$.

## 3 Computing Posteriors of Features

Every DAG must have a root node, that is, a node with no parents. Let $\mathcal{G}$ denote the set of all DAGs over $V$, and for any $S \subseteq V$ let $\mathcal{G}^+(S)$ be the set of DAGs over $V$ such that all variables in $V - S$ are root nodes (setting $\mathcal{G}^+(V) = \mathcal{G}$). Since every DAG must have a root node we have $\mathcal{G} = \cup_{j \in V}\mathcal{G}^+(V - \{j\})$. We can compute the summation over all the possible DAGs in Eq. (7) by summing over the DAGs in $\mathcal{G}^+(V - \{j\})$ separately. However there are overlaps between the set of graphs in $\mathcal{G}^+(V - \{j\})$, and in fact $\cap_{j \in T}\mathcal{G}^+(V - \{j\}) = \mathcal{G}^+(V - T)$. We can correct for those overlaps using the inclusion-exclusion principle. Define the following function for all $S \subseteq V$

$$RR(S) \equiv \sum_{G \in \mathcal{G}^+(S)} \prod_{i \in S} B_i(Pa_i). \quad (9)$$

We have $P(f, D) = RR(V)$ since $\mathcal{G}^+(V) = \mathcal{G}$. Then by the weighted inclusion-exclusion principle, Eq. (7) becomes

$$P(f, D) = RR(V) = \sum_{G \in \mathcal{G}} \prod_{i=1}^n B_i(Pa_i)$$

$$= \sum_{k=1}^{|V|}(-1)^{k+1} \sum_{\substack{T \subseteq V \\ |T|=k}} \sum_{G \in \mathcal{G}^+(V-T)} \prod_{i=1}^n B_i(Pa_i)$$

$$= \sum_{k=1}^{|V|}(-1)^{k+1} \sum_{\substack{T \subseteq V \\ |T|=k}} \prod_{j \in T} B_j(\emptyset) \sum_{G \in \mathcal{G}^+(V-T)} \prod_{i \in V-T} B_i(Pa_i)$$

$$= \sum_{k=1}^{|V|}(-1)^{k+1} \sum_{\substack{T \subseteq V \\ |T|=k}} \prod_{j \in T} B_j(\emptyset) RR(V - T), \quad (10)$$

which says that $P(f, D)$ can be computed in terms of $RR(S)$. Next we show that $RR(S)$ for all $S \subseteq V$ can be computed recursively. We define the following function for each $i \in V$ and all $S \subseteq V - \{i\}$

$$A_i(S) \equiv \sum_{Pa_i \subseteq S} B_i(Pa_i), \quad (11)$$

where in particular $A_i(\emptyset) = B_i(\emptyset)$. We have the following results, which roughly correspond to the backward computation in [Koivisto, 2006].

**Proposition 1**

$$P(f, D) = RR(V), \quad (12)$$

where $RR(S)$ can be computed recursively by

$$RR(S) = \sum_{k=1}^{|S|}(-1)^{k+1} \sum_{\substack{T \subseteq S \\ |T|=k}} RR(S - T) \prod_{j \in T} A_j(V - S) \quad (13)$$

with the base case $RR(\emptyset) = 1$.

*Proof:* We will say a node $j \in S$ is a *source* node in $G \in \mathcal{G}^+(S)$ if none of $j$'s parents are in $S$, that is, $Pa_j \cap S = \emptyset$. For $T \subseteq S$ let $\mathcal{G}^+(S, T)$ denote the set of DAGs in $\mathcal{G}^+(S)$ such that all the variables in $T$ are source nodes (setting $\mathcal{G}^+(S, \emptyset) = \mathcal{G}^+(S)$). It is clear that $\mathcal{G}^+(S) = \cup_{j \in S}\mathcal{G}^+(S, \{j\})$, and that $\cap_{j \in T}\mathcal{G}^+(S, \{j\}) = \mathcal{G}^+(S, T)$. The summation over the DAGs in $\mathcal{G}^+(S)$ in Eq. (9) can be computed by summing over the DAGs in $\mathcal{G}^+(S, \{j\})$ separately and correcting for the overlapped DAGs. Define

$$RF(S, T) \equiv \sum_{G \in \mathcal{G}^+(S,T)} \prod_{i \in S} B_i(Pa_i), \quad \text{for any } T \subseteq S, \quad (14)$$

where $RF(S, \emptyset) = RR(S)$. Then by the weighted inclusion-exclusion principle, $RR(S)$ in Eq. (9) can be computed as

$$RR(S) = \sum_{k=1}^{|S|}(-1)^{k+1} \sum_{T \subseteq S, |T|=k} RF(S, T). \quad (15)$$

$RR(S)$ and $RF(S, T)$ can be computed recursively as follows. For $|T| = 1$,

$$RF(S, \{j\}) = \sum_{G \in \mathcal{G}^+(S,\{j\})} B_j(Pa_j) \prod_{i \in S-\{j\}} B_i(Pa_i)$$

$$= [\sum_{Pa_j \subseteq V-S} B_j(Pa_j)][\sum_{G \in \mathcal{G}^+(S-\{j\})} \prod_{i \in S-\{j\}} B_i(Pa_i)]$$

(see Figure 1(a))

$$= A_j(V - S)RR(S - \{j\}). \quad (16)$$

Similarly, for any $T \subseteq S$

$$RF(S, T) = \sum_{G \in \mathcal{G}^+(S,T)} \prod_{j \in T} B_j(Pa_j) \prod_{i \in S-T} B_i(Pa_i)$$

$$= \prod_{j \in T}[\sum_{Pa_j \subseteq V-S} B_j(Pa_j)][\sum_{G \in \mathcal{G}^+(S-T)} \prod_{i \in S-T} B_i(Pa_i)]$$

(see Figure 1(b))

$$= \prod_{j \in T} A_j(V - S) \cdot RR(S - T). \quad (17)$$

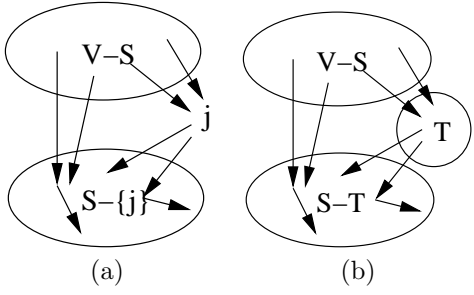

(a) (b)

Figure 1: Figures illustrating the proof of Proposition 1.

Combing Eqs. (15) and (17) we obtain Eq. (13).  □

Based on Proposition 1, provided that the functions $A_j$ have been computed, $P(f, D)$ can be computed in the manner of dynamic programming, starting from the base case $RR(\emptyset) = 1$, then $RR(\{j\}) = A_j(V - \{j\})$, and so on, until $RR(V)$.

Given the functions $B_i$, the functions $A_i$ as defined in Eq. (11) can be computed using the fast Möbius transform algorithm in time $O(n2^n)$ (for a fixed $i$) [Kennes and Smets, 1990]. In the case of a fixed indegree bound $k$, $B_i(Pa_i)$ is zero when $Pa_i$ contains more than $k$ elements. Then $A_i(S)$ for all $S \subseteq V - \{i\}$ can be computed more efficiently using the truncated Möbius transform algorithm given in [Koivisto and Sood, 2004] in time $O(k2^n)$ (for a fixed $i$).

The functions $RR$ may be computed more efficiently if we precompute the product of $A_j$. Define

$$AA(S, T) \equiv \prod_{j \in T} A_j(V - S) \quad \text{for } T \subseteq S \subseteq V. \quad (18)$$

Then using Eq. (18) for $AA(S, T - \{j\})$ we have

$$AA(S, T) = A_j(V - S)AA(S, T - \{j\}) \quad \text{for any } j \in T. \quad (19)$$

For a fixed $S$, $AA(S, T)$ for all $T \subseteq S$ can be computed in the manner of dynamic programming in $O(2^{|S|})$ time starting with $AA(S, \{j\}) = A_j(V - S)$. We then compute $RR(S)$ using

$$RR(S) = \sum_{k=1}^{|S|} (-1)^{k+1} \sum_{\substack{T \subseteq S \\ |T|=k}} RR(S - T) AA(S, T) \quad (20)$$

in $O(2^{|S|})$ time. The functions $RR(S)$ for all $S \subseteq V$ can be computed in $\sum_{k=0}^{n} \binom{n}{k} 2^k = 3^n$ time.

In summary, we obtain the following results.

**Theorem 1** *Given a fixed maximum indegree $k$, $P(f|D)$ can be computed in $O(3^n + kn2^n)$ time.*

## 4 Computing Posterior Probabilities for All Edges

If we want to compute the posterior probabilities of all $O(n^2)$ potential edges, we can compute $RR(V)$ for each edge separately and solve the problem in $O(n^2 3^n)$ total time. However, there is a large overlapping in the computations for different edges. After computing $P(D)$ with constant features $f_i(Pa_i) = 1$ for all $i \in V$, the computation for an edge $i \rightarrow j$ only needs to change the function $f_j$ and therefore $B_j$ and $A_j$. We can take advantage of this overlap and reduce the total time for computing for all edges.

Inspired by the forward-backward algorithm in [Koivisto, 2006], we developed an algorithm that can compute all edge posterior probabilities in $O(n3^n)$ total time. The computations of $P(f, D)$ in Section 3 are based exploiting root nodes and roughly correspond to the backward computation in [Koivisto, 2006]. Next we first describe a computation of $P(f, D)$ by exploiting sink nodes (nodes that have no outgoing edges) which roughly corresponds to the computation in [Koivisto and Sood, 2004] called forward computation in [Koivisto, 2006]. Then we describe how to combine the two computations to reduce the total computation time.

### 4.1 Computing $P(f, D)$ by exploiting sinks

For any $S \subseteq V$, let $\mathcal{G}(S)$ denote the set of all the possible DAGs over $S$, and $\mathcal{G}(S, T)$ denote the set of all the possible DAGs over $S$ such that all the variables in $T \subseteq S$ are sinks. We have $\mathcal{G}(V) = \mathcal{G}$ and $\mathcal{G}(S, \emptyset) = \mathcal{G}(S)$. For any $S \subseteq V$ define

$$H(S) \equiv \sum_{G \in \mathcal{G}(S)} \prod_{i \in S} B_i(Pa_i). \quad (21)$$

We have $P(f, D) = H(V)$ since $\mathcal{G}(V) = \mathcal{G}$. As in Section 3 we can show that $H(S)$ can be computed recursively. For any $T \subseteq S \subseteq V$, define

$$F(S, T) \equiv \sum_{G \in \mathcal{G}(S,T)} \prod_{i \in S} B_i(Pa_i), \quad (22)$$

where $F(S, \emptyset) = H(S)$. We have the following results.

**Proposition 2**

$$P(f, D) = H(V), \quad (23)$$

and $H(S)$ can be computed recursively by

$$H(S) = \sum_{k=1}^{|S|}(-1)^{k+1}\sum_{\substack{T\subseteq S \\ |T|=k}} H(S-T)\prod_{j\in T} A_j(S-T) \quad (24)$$

with the base case $H(\emptyset) = 1$.

*Proof:* Since every DAG has a sink we have $\mathcal{G}(S) = \cup_{j\in S}\mathcal{G}(S,\{j\})$. It is clear that $\cap_{j\in T}\mathcal{G}(S,\{j\}) = \mathcal{G}(S,T)$. The summation over the DAGs in $\mathcal{G}(S)$ in Eq. (21) can be computed by summing over the DAGs in $\mathcal{G}(S,\{j\})$ separately and correcting for the overlapped DAGs. By the weighted inclusion-exclusion principle, $H(S)$ in Eq. (21) can be computed as

$$H(S) = \sum_{k=1}^{|S|}(-1)^{k+1}\sum_{T\subseteq S,|T|=k} F(S,T). \quad (25)$$

$H(S)$ and $F(S,T)$ can be computed recursively as follows. For $|T| = 1$, we have

$$F(S,\{j\}) = \sum_{G\in\mathcal{G}(S,\{j\})}\prod_{i\in S} B_i(Pa_i)$$
$$= [\sum_{Pa_j\subseteq S-\{j\}} B_j(Pa_j)][\sum_{G\in\mathcal{G}(S-\{j\})}\prod_{i\in S-\{j\}} B_i(Pa_i)]$$
(see Figure 2(a))
$$= A_j(S-\{j\})H(S-\{j\}). \quad (26)$$

Similarly, for any $j \in T \subseteq S$

$$F(S,T) = \sum_{G\in\mathcal{G}(S,T)}\prod_{i\in S} B_i(Pa_i)$$
$$= [\sum_{Pa_j\subseteq S-T} B_j(Pa_j)][\sum_{G\in\mathcal{G}(S-\{j\},T-\{j\})}\prod_{i\in S-\{j\}} B_i(Pa_i)]$$
(see Figure 2(b))
$$= A_j(S-T)F(S-\{j\},T-\{j\}). \quad (27)$$

Let $T = \{j_1,\ldots,j_k\}$. Repeatedly applying Eq. (27) and using the fact $(S-T')-(T-T') = S-T$ for any $T' \subseteq T$, we obtain

$$F(S,T) = A_{j_1}(S-T)A_{j_2}(S-T)F(S-\{j_1,j_2\},T-\{j_1,j_2\})$$
$$= \ldots$$
$$= A_{j_1}(S-T)\cdots A_{j_{k-1}}(S-T)F(S-\{j_1,\ldots,j_{k-1}\},\{j_k\})$$
$$= H(S-T)\prod_{j\in T} A_j(S-T), \quad (28)$$

where Eq. (26) is applied in the last step. Finally combining Eqs. (28) and (25) leads to (24). □

Based on Proposition 2, $H(S)$ can be computed in the manner of dynamic programming. Each $H(S)$ is computed in time $\sum_{k=1}^{|S|}\binom{|S|}{k}k = |S|2^{|S|-1}$. All $H(S)$ for $S \subseteq V$ can be computed in time $\sum_{k=1}^{n}\binom{n}{k}k2^{k-1} = n3^{n-1}$. We could store all $F(S,T)$ and compute all $H(S)$ in time $O(3^n)$. But the memory requirement would become $O(4^n)$ instead of $O(n2^n)$.

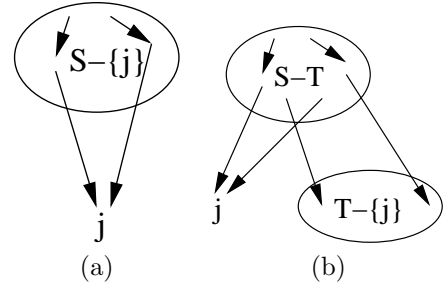

Figure 2: Figures illustrating the proof of Proposition 2.

We could compute the posterior of a feature using $P(f,D) = H(V)$ but this is a factor of $n$ slower than computing $RR(V)$. Next we show how to reduce the total time for computing the posterior probabilities of all edges by combining the contributions of $H(S)$ and $RR(S)$.

### 4.2 Computing posteriors for all edges

Consider the summation over all the possible DAGs in Eq. (7). Assume that we are interested in computing the posterior probability of an edge $i \to v$. We want to extract the contribution of $B_v$ from the rest of $B_i$. The idea is as follows. For a fixed node $v$, we can break a DAG into the set of ancestors $U$ of $v$ and the set of nonancestors $V - \{v\} - U$.[1] Roughly speaking, conditioned on $U$, the summation over all DAGs can be decomposed into the contributions from the summation over DAGs over $U$, which corresponds to the computation of $H(U)$, and the contributions from the summation over DAGs over $V - \{v\} - U$ with the variables in $U \cup \{v\}$ as root nodes, which corresponds to the computation of $RR(V - \{v\} - U)$.

Define, for any $v \in V$ and $U \subseteq V - \{v\}$ the following function

$$K_v(U) \equiv \sum_{T\subseteq V-\{v\}-U}(-1)^{|T|}RR(V-\{v\}-U-T)\prod_{j\in T} A_j(U). \quad (29)$$

We have the following results.

---
[1] Or we can break a DAG into the set of nondescendants $U$ of $v$ and the set of descendants $V - \{v\} - U$. It could be shown that we can also use this way of breaking DAGs to derive Proposition 3. But this is not exploited in the paper.

**Proposition 3**

$$P(f, D) = \sum_{U \subseteq V - \{v\}} A_v(U) H(U) K_v(U). \quad (30)$$

The proof of Proposition 3 is given in the Appendix.

Note that in Eq. (30) the computations of $H(U)$ and $K_v(U)$ do not rely on $B_v$ and all the contribution from $B_v$ to $P(f, D)$ is represented by $A_v$. After we have computed the functions $A_v$, $H$ and $K_v$, $P(f, D)$ can be computed using Eq. (30) in time $O(2^n)$.

To compute $K_v(U)$, we can precompute the product of $A_j$. Define

$$\eta(U, T) \equiv \prod_{j \in T} A_j(U). \quad (31)$$

Then $K_v(U)$ can be computed as

$$K_v(U) = \sum_{T \subseteq V - \{v\} - U} (-1)^{|T|} RR(V - \{v\} - U - T) \eta(U, T), \quad (32)$$

where $\eta(U, T)$ can be computed recursively as

$$\eta(U, T) = A_j(U) \eta(U, T - \{j\}) \quad \text{for any } j \in T. \quad (33)$$

For a fixed $U$, all $\eta(U, T)$ for $T \subseteq V - \{v\} - U$ can be computed in $2^{n-1-|U|}$ time, and then $K_v(U)$ can be computed in $2^{n-1-|U|}$ time. For a fixed $v$ all $K_v(U)$ for $U \subseteq V - \{v\}$ can be computed in $\sum_{k=0}^{n-1} \binom{n-1}{k} 2^{n-1-k} = 3^{n-1}$ time.

Based on Proposition 3, to compute the posterior probabilities for all possible edges, we can use the following algorithm.

1. Precomputation. Set constant feature $f(G) \equiv 1$. Compute functions $B_i$, $A_i$, $RR$, $H$, and $K_i$.

2. For each edge $u \to v$:

   (a) For all $S \subseteq V - \{v\}$, recompute $A_v(S)$.
   (b) Compute $P(f, D)$ using Eq. (30). Then $P(f|D) = P(f, D)/RR(V)$.

For a fixed maximum indegree $k$, Step 1 takes time $O(n3^n)$ as discussed before, and Step 2 takes time $O(n^2(k2^n + 2^n))$. It takes $O(n3^n + kn^2 2^n)$ total time to compute the posterior probabilities for all possible edges.

The computation time of Step 2 can be further reduced by a factor of $n$ using the techniques described in [Koivisto, 2006]. Plug in the definition of $A_v(U)$

into Eq. (30)

$$P(f, D) = \sum_{U \subseteq V - \{v\}} [\sum_{Pa_v \subseteq U} B_v(Pa_v)] H(U) K_v(U)$$

$$= \sum_{Pa_v \subseteq V - \{v\}} B_v(Pa_v) \sum_{Pa_v \subseteq U \subseteq V - \{v\}} H(U) K_v(U)$$

$$= \sum_{Pa_v \subseteq V - \{v\}} B_v(Pa_v) \Gamma_v(Pa_v), \quad (34)$$

where for all $Pa_v \subseteq V - \{v\}$ we define

$$\Gamma_v(Pa_v) \equiv \sum_{Pa_v \subseteq U \subseteq V - \{v\}} H(U) K_v(U). \quad (35)$$

For a fixed maximum indegree $k$, since we set $B_v(Pa_v)$ to be zero for $Pa_v$ containing more than $k$ variables we need to compute the function $\Gamma_v(Pa_v)$ only at sets $Pa_v$ containing at most $k$ elements, which can be computed using the $k$-truncated downward Möbius transform algorithm described in [Koivisto, 2006] in $O(k2^n)$ time for all $Pa_v$ (for a fixed $v$). Then $P(f, D)$ for an edge $u \to v$ can be computed as

$$P(u \to v, D) = \sum_{\substack{u \in Pa_v \subseteq V - \{v\} \\ |Pa_v| \leq k}} B_v(Pa_v) \Gamma_v(Pa_v), \quad (36)$$

which takes $O(n^k)$ time.

In summary, we propose the algorithm in Figure 3 to compute the posterior probabilities for all possible edges. The main result of the paper is summarized in the following theorem.

**Theorem 2** *For a fixed maximum indegree $k$, the posterior probabilities for all $n(n-1)$ possible edges can be computed in $O(n3^n)$ total time.*

## 5 Experimental Results

We have implemented the algorithm in Figure 3 in the C++ language and run some experiments to demonstrate its capabilities. We compared our implementation with REBEL[2], a C++ language implementation of the DP algorithm in [Koivisto, 2006].

We used BDe score for $score_i(Pa_i : D)$ (with the hyperparameters $\alpha_{x_i;pa_i} = 1/(|Dm(X_i)| \cdot |Dm(Pa_i)|)$) [Heckerman et al., 1995]. In the experiments our algorithm used a uniform structure prior $P(G) = 1$ and REBEL used a structure prior specified in [Koivisto, 2006]. All the experiments were run under Linux on an ordinary desktop PC with a 3.0GHz Intel Pentium processor and 2.0GB of memory.

**Algorithm** Computing posteriors of all edges given maximum indegree $k$

1. Precomputation. Set trivial feature $f(G) \equiv 1$

   (a) For all $i \in V$, $Pa_i \subseteq V - \{i\}$ with $|Pa_i| \leq k$, compute $B_i(Pa_i)$. Time complexity $O(n^{k+1})$.
   
   (b) For all $i \in V$, $S \subseteq V - \{i\}$, compute $A_i(S)$. Time complexity $O(kn2^n)$.
   
   (c) For all $S \subseteq V$, compute $RR(S)$. Time complexity $O(3^n)$.
   
   (d) For all $S \subseteq V$, compute $H(S)$. Time complexity $O(n3^n)$.
   
   (e) For all $i \in V$, $S \subseteq V - \{i\}$, compute $K_i(S)$. Time complexity $O(n3^n)$.
   
   (f) For all $i \in V$, $Pa_i \subseteq V - \{i\}$ with $|Pa_i| \leq k$, compute $\Gamma_i(Pa_i)$. Time complexity $O(kn2^n)$.

2. For each edge $u \to v$, compute $P(u \to v|D)$ using Eq. (36), and output $P(u \to v|D) = P(u \to v, D)/RR(V)$. Time complexity $O(n^{k+2})$.

Figure 3: Algorithm for computing the posterior probabilities for all possible edges in time complexity $O(n3^n)$ assuming a fixed maximum indegree $k$.

Table 1: The speed of our algorithm (in second).

| Name      | $n$ | $m$ | $k$ | $T_B$  | Ours   | REBEL  |
|-----------|-----|-----|-----|--------|--------|--------|
| Iris      | 5   | 150 | 4   | 2.2e-3 | 3.5e-3 | 3.1e-3 |
| TicTacToe | 10  | 958 | 4   | 4.7e-1 | 6.2e-1 | 5.1e-1 |
|           |     |     | 5   | 9.1e-1 | 1.1    | 9.4e-1 |
|           |     |     | 6   | 1.3    | 1.5    | 1.4    |
|           |     |     | 9   | 1.7    | 1.9    | 1.7    |
| Zoo       | 17  | 101 | 4   | 1.4    | 602.3  | 13.4   |
|           |     |     | 5   | 4.4    | 607.0  | 19.2   |
|           |     |     | 6   | 11.5   | 610.6  | 28.7   |
| Synthetic | 20  | 500 | 4   | 9.2    | 23083  | 128.3  |

## 5.1 Speed Test

We tested our algorithm on several data sets from the UCI Machine Learning Repository [Asuncion and Newman, 2007]: Iris, Tic-Tac-Toe, and Zoo. All the data sets contain discrete variables (or are discretized) and have no missing values. We also tested our algorithm on a synthetic data set coming with REBEL. For each data set, we ran our algorithm and REBEL to compute the posterior probabilities for all potential edges. The time taken under different maximum indegree $k$ is reported in Table 1, which also lists the number of variables $n$ and the number of instances $m$ for each data set. We also show the time $T_B$ for computing local scores in Table 1 as this time also depends on the number of instances $m$ in a data set.

The results demonstrate that our algorithm is capable of computing the posterior probabilities for all potential edges in networks over around $n = 20$ variables. The memory requirement of the algorithm is $O(n2^n)$, the same as REBEL, which will limit the use of the algorithm to about $n = 25$ variables. It may take our current implementation a few months for $n = 25$.

## 5.2 Comparison of computations

For the Tic-Tac-Toe data set with $n = 10$, our algorithm is capable of computing the "true" exact edge posterior probabilities by setting the maximum indegree $k = 9$,[3] although an exhaustive enumeration of DAGs with $n = 10$ would not be feasible. We then vary the maximum indegree $k$ and compare the edge posterior probabilities computed by our algorithm with the true probabilities. The results are shown as scatter plots in Figure 4 (Note that in these graphs most of the points are located at (0,0) or closely nearby). Each point in a scatter plot corresponds to an edge with its $x$ and $y$ coordinates denoting the posterior computed by the two compared algorithms. We see that with the increase of $k$ the computed probabilities gradually approach the true probabilities. With $k = 3$ the computed probabilities already converge to the true probabilities. Studying the effects of the approximation due to the maximum indegree restriction in general need more substantial experiments and is beyond the scope of this paper.

We also compared the exact posterior probabilities computed by REBEL (setting $k = 9$) with the true probabilities. The results are shown in Figure 5. We see that the exact probabilities computed by REBEL without indegree bound sometimes still differ with the true probabilities. This is due to the highly non uniform structure prior used by REBEL.

---

[2]REBEL is available at http://www.cs.helsinki.fi/u/mkhkoivi/REBEL/.

[3]We will call the exact posterior probabilities computed using uniform structure prior $P(G) = 1$ the "true" probabilities.

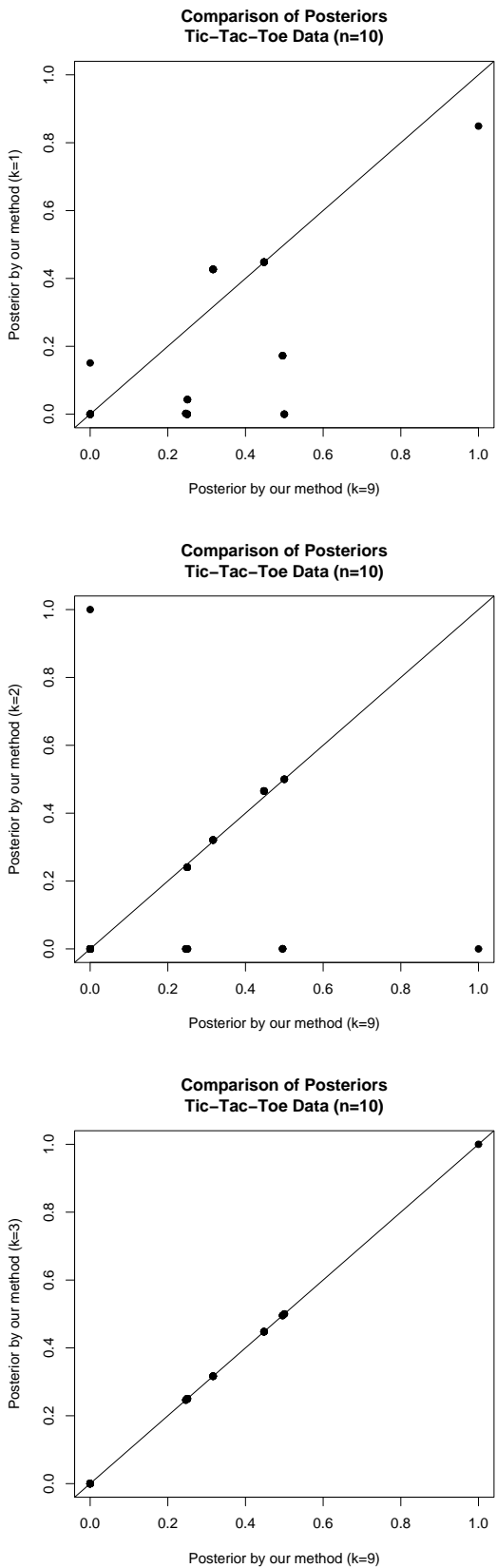

Figure 4: Scatter plots that compare posterior probability of edges on the Tic-Tac-Toe data set as computed by our algorithm with different $k$ against the true posterior.

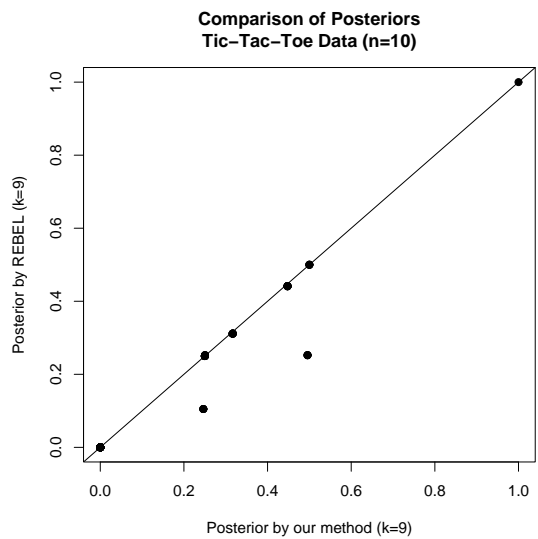

Figure 5: A scatter plot that compares posterior probability of edges on the Tic-Tac-Toe data set as computed by REBEL against the "true" posterior.

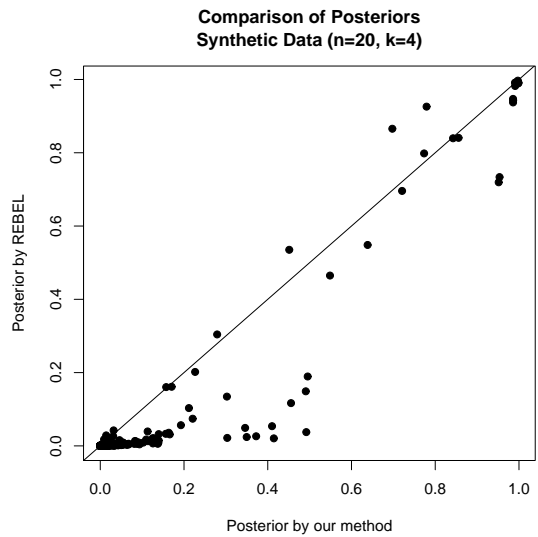

Figure 6: A scatter plot that compares posterior probability of edges on the Synthetic data set as computed by REBEL and our algorithm.

We compared our algorithm with REBEL over a larger network, the synthetic data set with $n = 20$. The results are shown in Figure 6. We see that with the same maximum indegree, the computed probabilities often differ. Again, this can be attributed to the non uniform structure prior used by REBEL.

## 6 Conclusion

We have presented an algorithm that can compute the exact marginal posterior probability of a single edge in $O(3^n)$ time and the posterior probabilities for all $n(n-1)$ potential edges in $O(n3^n)$ total time. We demonstrated its capability on data sets containing up to 20 variables.

The main advantage of our algorithm over the current state-of-the-art algorithms, the DP algorithm in [Koivisto, 2006] and the order MCMC in [Friedman and Koller, 2003], for computing the posterior probabilities of structural features is that those algorithms require special structure prior $P(G)$ that is highly non uniform while we allow general prior $P(G)$.

Our algorithm computes exact posterior probabilities and works in moderate size networks (about 20 variables), which make it a useful tool for studying several problems in learning Bayesian networks. One application is to assess the quality of the DP algorithm due to the influence of non uniform prior $P(G)$. Another application is to study the effects of the approximation due to the maximum indegree restriction. We have shown some initial experimental results in Section 5. Other potential applications include assessing the quality of approximate algorithms (e.g., MCMC algorithms), studying the effects of data sample size on the learning results, and studying the effects of model parameters (such as parameter priors) on the learning results.

**Acknowledgments**

This research was partly supported by NSF grant IIS-0347846.

## Appendix : Proof of Proposition 3

For a fixed node $v$, we can break a DAG uniquely into the set of ancestors $S$ of $v$ and the set of nonancestors $V - \{v\} - S$. For $v \notin S$ let $\mathcal{G}^v(S)$ denote the set of DAGs over $S \cup \{v\}$ such that every node in $S$ is an ancestor of $v$. Then the summation over all possible DAGs in Eq. (7) can be decomposed into

$$P(f, D) = \sum_{S \subseteq V-\{v\}} [\sum_{G \in \mathcal{G}^v(S)} \prod_{i \in S \cup \{v\}} B_i(Pa_i)]$$
$$\cdot [\sum_{G \in \mathcal{G}^+(V-\{v\}-S)} \prod_{i \in V-\{v\}-S} B_i(Pa_i)]$$
$$= \sum_{S \subseteq V-\{v\}} LL_v(S) RR(V - \{v\} - S), \quad (37)$$

where for any $S \subseteq V - \{v\}$ we define

$$LL_v(S) \equiv \sum_{G \in \mathcal{G}^v(S)} \prod_{i \in S \cup \{v\}} B_i(Pa_i). \quad (38)$$

$\mathcal{G}^v(S)$ consists of the set of DAGs over $S \cup \{v\}$ in which $v$ is the unique sink. We have

$$\mathcal{G}(S \cup \{v\}, \{v\}) = \mathcal{G}^v(S) \cup (\cup_{j \in S} \mathcal{G}(S \cup \{v\}, \{v, j\})), \quad (39)$$

from which, by the weighted inclusion-exclusion principle, we obtain

$$LL_v(S) = \sum_{G \in \mathcal{G}(S \cup \{v\}, \{v\})} \prod_{i \in S \cup \{v\}} B_i(Pa_i)$$
$$- \sum_{k=1}^{|S|} (-1)^{k+1} \sum_{\substack{T \subseteq S \\ |T|=k}} \sum_{G \in \mathcal{G}(S \cup \{v\}, T \cup \{v\})} \prod_{i \in S \cup \{v\}} B_i(Pa_i)$$
$$= F(S \cup \{v\}, \{v\}) - \sum_{k=1}^{|S|} (-1)^{k+1} \sum_{\substack{T \subseteq S \\ |T|=k}} F(S \cup \{v\}, T \cup \{v\})$$
$$= \sum_{T \subseteq S} (-1)^{|T|} F(S \cup \{v\}, T \cup \{v\})$$
$$= \sum_{T \subseteq S} (-1)^{|T|} A_v(S-T) F(S, T)$$
$$= \sum_{U \subseteq S} (-1)^{|U|+|S|} A_v(U) F(S, S-U). \quad (40)$$

Plugging Eq. (40) into Eq. (37), we obtain

$$P(f, D) = \sum_{S \subseteq V-\{v\}} \sum_{U \subseteq S} (-1)^{|U|+|S|} A_v(U)$$
$$\cdot F(S, S-U) RR(V - \{v\} - S)$$
$$= \sum_{U \subseteq V-\{v\}} \sum_{U \subseteq S \subseteq V-\{v\}} (-1)^{|U|+|S|} A_v(U)$$
$$\cdot F(S, S-U) RR(V - \{v\} - S)$$
$$= \sum_{U \subseteq V-\{v\}} A_v(U) H(U) \sum_{U \subseteq S \subseteq V-\{v\}} (-1)^{|U|+|S|}$$
$$\cdot \prod_{j \in S-U} A_j(U) RR(V - \{v\} - S)$$
$$= \sum_{U \subseteq V-\{v\}} A_v(U) H(U) K_v(U), \quad (41)$$

where we have used the definition of function $K_v(U)$ in Eq. (29).